# The Serendipity of Claude AI : Case of the 13 Low-Resource National Languages of Mali

**Authors:** *By Alou DEMBELE, Nouhoum Souleymane COULIBALY, Michael LEVENTHAL RobotsMali AI4D Lab, Bamako, Mali, research@robotsmali.org*

## Abstract

Recent advances in artificial intelligence (AI) and natural language processing (NLP) have improved the representation of underrepresented languages. However, most languages, including Mali's 13 official national languages, continue to be poorly supported or unsupported by automatic translation and generative AI. This situation appears to have slightly improved with certain recent LLM releases. The study evaluated Claude AI's translation performance on each of the 13 national languages of Mali. In addition to ChrF2 and BLEU scores, human evaluators assessed translation accuracy, contextual consistency, robustness to dialect variations, management of linguistic bias, adaptation to a limited corpus, and ease of understanding. The study found that Claude AI performs robustly for languages with very modest language resources and, while unable to produce understandable and coherent texts for Malian languages with minimal resources, still manages to produce results which demonstrate the ability to mimic some elements of the language.

## 1. Introduction

In recent years, advances in artificial intelligence (AI) and natural language processing (NLP) have marked a significant advance in the representation and development of underrepresented languages. However, most of the world's languages, often referred to as "low-resource languages", still remain either not supported or insufficiently supported due to the limited availability of data and language resources, and market, economic, and global inequality factors.

Mali, a multilingual country with 13 official languages, including Bamanankan (Bambara), Bomu, Bozo, Dɔgɔsɔ (Dogon), Fulfulde (Fula), Hassaniya Arabic, Mamara (Minyanka), Maninka, Soninke, Sɔõɔy (Songhay), Senara, Tàmàsàyt (Tamasheq) and Xaasongaxanno (Kassonke), faces severe challenges in digital inclusion limiting economic development, educational advancement, and preservation of cultural heritage (Bird, 2020 ; Nekoto et al., 2020). These languages share in common a penury of language resources needed to train AI and NLP systems which could play a role in lessening the digital divide (Hammarström et al., 2018). This penury extends from severe in the case of a language like Bambara which has very limited resources to catastrophic for languages like Bomu and Bozo with an almost complete absence of language resources.

The need for innovative methods for low-resource languages has spawned varied strategies, such as transfer learning, zero-shot learning, and pre-trained models in related languages (Ruder, 2021; Adelani et al., 2022 ). Claude AI, like other advanced models, uses cross-linguistic transfer to fill the resource gap by applying knowledge from better-resourced languages to lesser-known languages (Conneau et al., 2020). Although promising, these methodologies may still be challenged by languages whose typological and structural characteristics differ greatly from those of more resource-rich languages (Pires et al., 2019).

By examining the effectiveness of Claude AI in supporting Mali's national languages, this research highlights the challenges and opportunities of AI in linguistic inclusion. Improving AI tools for low-resource languages could provide Mali's diverse linguistic communities with greater access to technologies, thereby promoting greater educational, sociocultural, and economic inclusion.

Our evaluations of Claude's performance were carried out despite the fact that its use in Mali has been geo-blocked. Our researchers circumvented this restriction using a VPN and phone numbers in unblocked countries. Despite the exclusion of Malian users, Claude AI manages to translate certain languages of Mali without having received specific training for them. It can understand and produce translations in less-represented languages, including several Malian languages, exploiting knowledge of common linguistic structures and contexts.

2. Literature review

Most AI and NLP models still struggle with low-resource languages, which suffer from a shortage of digitized data, annotated corpora, and linguistic resources *(Joshi et al., 2020)*.

Claude AI's capabilities in translating low-resource languages, particularly for African languages, have attracted attention in recent studies, although challenges remain. According to Enis and Hopkins (2024), Claude 3 Opus shows strong proficiency when translating low-resource languages into English, outperforming Google Translate and Meta's NLLB-54B on benchmark datasets such as FLORES-200. They noted, however, that "Claude performs significantly better when translating into English than from English," a limitation affecting African languages that are less widely studied in machine translation (MT) research (Enis & Hopkins , 2024)

Enis and Hopkins (2024) showed that Claude outperformed these models in 25% of the languages evaluated, although its performance varied depending on the translation direction and the dataset used.

In a survey study, Sunyu Transphere (2024) examined Claude's handling of various languages, including some from Africa. They reported that Claude's translation quality varies significantly depending on the dataset and translation direction, noting that "his abilities are significantly lower when translating from English to low-resource languages ", reflecting a typical limitation in many LLMs (Sunyu Transphere, 2024)

Enis and Hopkins (2024) argue for the expansion of high-quality linguistic resources for African languages, a move that could help refine Claude's abilities for Mali's official languages. Currently, they suggest that the application of Claude in these contexts is better suited to translation into English rather than from English, where it continues to face accuracy issues (Enis & Hopkins, 2024; Sunyu Transphere, 2024).

*The challenge of low-resource languages in NLP*

African languages, particularly those of Mali, present an additional layer of complexity as they often feature dialectal variations, complex phonetic and grammatical systems, and limited historical documentation *(Hammarström et al., 2017; Orife et al., 2020)*. The development of precise NLP models for these languages has therefore become an important area of research in recent years. *(Ruder et al., 2021)*.

*Claude AI's approach to low-resource languages*

Claude AI is trained to handle numerous languages, leveraging cross-linguistic transfer learning and other cutting-edge techniques. However, models like Claude AI are typically trained on large, multilingual datasets that heavily favor languages with a significant digital presence, leaving low-resource languages with limited support. *(Smith et al., 2022)*. The effectiveness of Claude AI on languages such as Bambara and Songhay is further complicated by their phonological diversity. Cross-linguistic transfer, which involves applying knowledge from resource-rich languages to low-resource languages, can be effective, but often does not take into account the linguistic nuances essential for good comprehension and translation *(Conneau et al., 2020; Pires et al., 2019)*.

*Challenges specific to the languages of Mali*

Mali's national languages embody diverse linguistic structures and dialect variations, making data collection and model training particularly difficult. Languages like Dɔgɔsɔ (Dogon) or Senara, for example, contain significant regional variations, making it difficult to create unified datasets *(Hammarström et al., 2017)*. Additionally, the languages of Mali are predominantly oral, with few printed works and fewer still in electronic form, restricting the corpora available for model training. *(Bird, 2020)*. Malian languages are weakly supported by systems of standardized spelling and linguistic resources such as dictionaries, making it difficult to pre-train and evaluate models with consistent data *(Joshi et al., 2020)*.

Bambara, among the Malian languages, is the best resourced, with a standardized orthography and an online dictionary adhering to this standard, an 11 million word electronic corpus hosted by INALCO, and a curated machine learning dataset of approximately 50,000 aligned French-Bambara sentences created by RobotsMali. *(Tapo et. al., 2020)*.

3. Methodology

A comprehensive suite of human and machine tests was performed for each language.

- Translation of text
- Comparison to reference texts
- Automated scoring (BLEU and chrF)

*Claude AI performance evaluation criteria for low-resource languages in Mali*

To effectively evaluate Claude AI on low-resource Malian languages, the criteria was developed that match the linguistic and technical particularities of these languages in order to clearly understand its performance and identify avenues for improvement.

**Translation accuracy:** Evaluate the fidelity of the translations produced by Claude, taking into account the lexical, syntactic and semantic particularities specific to the languages of Mali, in particular their ability to reproduce specific idiomatic and cultural expressions, including those from linguistically isolated communities (e.g. Dogon, Tamasheq).

**Contextual consistency:** Measure the ability of AI to maintain consistency in texts where dialect variations are present, such as those observed in Senara and Dogon, especially for complex local language text generation tasks.

**Robustness in the face of dialect variations** : Observe whether Claude correctly manages the subtle differences between sub-dialects of languages like Dogon and Songhay, maintaining precision and consistency despite non-standardized spelling variations.

**Management of linguistic biases:** Identify and limit Claude's potential biases which favor dominant languages (French, English) in order to ensure faithful and uninfluenced representation of minority languages such as Mamara and Khassonke.

**Adaptation to a limited corpus** : Evaluate Claude's performance with little training data, a crucial requirement for languages underrepresented in digital resources (e.g. Senara, Tamasheq).

**Ease of understanding and use:** Test the fluency and understanding of the output generated by Claude with native speakers and domain specialists, particularly for practical applications such as the translation of administrative documents.

*Data sources and collection*

The 13 official languages of Mali, including Bambara, Fula, Soninke, and others, are poorly equipped with digital corpora. Collecting data for these languages required varied sources to assess Claude's performance on these low-resource languages.

*Geographic and demographic context*

Located in West Africa, Mali is a country of great linguistic diversity with approximately 22 million inhabitants, distributed between urban areas such as Bamako and rural areas. The 13 official national languages, alongside French, are a reflection of this cultural diversity and constitute a relevant basis for the evaluation of AI in this context.

*Evaluators*

There are two official, governmental organizations in Mali that serve as the ultimate authorities on each of Mali's 13 national languages.

- **Malian National Directorate of Non-Formal Education and National Languages** (DNENF-LN)

  The DNENF-LN focuses, essentially, on the use of Mali's national languages in an educational context. Since formal education is currently in French, the Directorate concerns itself with non-formal education, that is, education outside the K-12 classroom setting. There is one language unit for each national language of Mali.

- **Malian National Academy of Languages** (AMALAN)

  AMALAN is an, essentially, linguistic body responsible for the standardization and development of each national language. There is a linguistic unit for each national language of Mali.

*Materials and resources used*

- **Linguistic corpora** : Texts written in each of the thirteen spoken languages and French, extracted from the Malian Mining Code and from the story "Niamoye" with parallel translations in 12 of the 13 national languages of Mali and French.

4. **Presentation and analysis of results**

Automatic metrics, BLEU and ChrF++, could be applied to these texts and proved largely meaningful for quick, preliminary evaluation, particularly for comparing supported languages with significant resources. However, human evaluation, on both the materials with reference translations and on other collected text samples, was used to judge fluency, fidelity to meaning and cultural relevance. This type of evaluation provided the only meaningful feedback for the most resource-poor languages.

*Table 1: Claude's ChrF2 and BLEU scores when translating from French to each of the languages of Mali*
*Test 1: Text de Niamoye*

|    | Language  | Score ChrF2 | Score BLEU |
|----|-----------|-------------|------------|
| 0  | bambara   | 64.787      | 43.143     |
| 1  | bomu      | 13.572      | 0.763      |
| 2  | bozo      | 14.799      | 1.457      |
| 3  | dogon     | 17.409      | 1.901      |
| 4  | kassonke  | 29.716      | 5.122      |
| 5  | maninka   | 15.651      | 1.626      |
| 6  | minyanka  | 11.889      | 1.269      |
| 7  | hassaniya | 0.0         | 0.0        |
| 8  | fula      | 25.168      | 2.013      |
| 9  | senara    | 21.759      | 1.315      |
| 10 | songhay   | 21.204      | 1.943      |
| 11 | soninke   | 32.29       | 3.174      |
| 12 | tamasheq  | 22.706      | 1.977      |

*Table 2: Claude's ChrF2 and BLEU scores when translating from French into each of the languages of Mali*
*Case 2: Text of Mali's Mining Code*

|    | Language  | Score ChrF2 | Score BLEU |
|----|-----------|-------------|------------|
| 0  | bambara   | 59.841      | 38.466     |
| 1  | bomu      | 0.0         | 0.0        |
| 2  | bozo      | 19.153      | 0.925      |
| 3  | dogon     | 0.0         | 0.0        |
| 4  | kassonke  | 22.325      | 5.894      |
| 5  | maninka   | 21.051      | 2.61       |
| 6  | minyanka  | 0.0         | 0.0        |
| 7  | hassaniya | 14.127      | 4.852      |
| 8  | fula      | 0.0         | 0.0        |
| 9  | senara    | 0.0         | 0.0        |
| 10 | songhay   | 24.029      | 2.138      |
| 11 | soninke   | 29.407      | 8.528      |
| 12 | tamasheq  | 18.54       | 0.725      |

*Table 3: Evaluation by experts of AMALAN and the DNENF-LN Text of Niamoye*

| Language | Accuracy of the translation | | Fluidity of the text | | Orthography and grammar | |
|---|---|---|---|---|---|---|
| | AMALAN | DNENF-LN | AMALAN | DNENF-LN | AMALAN | DNENF-LN |
| Bambara | Good | Excellent | Fairly Fluid | Fairly Fluid | Excellent | Excellent |
| Fula | Very Good | Very Good | Somewhat Fluid | Somewhat Fluid | Good | Good |
| Soninke | Excellent | Good | Somewhat Fluid | Fairly Fluid | Good | Good |
| Songhay | Fair | Good | Not Fluid | Not Fluid | Good | Good |
| Tamasheq | Unusable | Unusable | Not Fluid | Somewhat Fluid | Very Good | Unusable |
| Dogon | Fair | Unusable | Not Fluid | Not Fluid | Very Poor | Unusable |
| Bozo | Very Poor | Fair | Not Fluid | Not Fluid | Poor | Unusable |
| Maninka | | | | Not Fluid | Good | Very Good |
| Kassonke | Poor | Very Poor | Somewhat Fluid | Not Fluid | Good | Unusable |
| Bomu | Fair | Unusable | Not Fluid | Not Fluid | Poor | Unusable |
| Minyanka | Very Poor | Fair | Somewhat Fluid | Somewhat Fluid | Unusable | Unusable |
| Senara | Unusable | Poor | Not Fluid | Not Fluid | Unusable | Unusable |
| Hassaniya | Fair | Unusable | Not Fluid | Not Fluid | Very Poor | Unusable |

*4 languages out of the 13 were considered good by human evaluators, namely Bambara, Fula, Soninke and Songhay*

**General Summary:**

The analysis aims to assess the quality of translations of texts from French into the thirteen Malian languages, taking into account fidelity to the source text, fluidity, coherence and linguistic adequacy.

The reference texts were created and validated for correctness by human experts in French and all Malian languages, with the exception of Hassaniya Arabic for Niamoye. The French text was translated by Claude AI into each Malian language and ChrF2 and BLEU scores were measured against the reference translations as presented in Tables 1 and 2.

Bambara has, by a considerable margin, the highest quantitative scores (64.8, 59.8 for ChrF2 and 43.1, 38.5 for BLEU for the two texts). Kassonke, and Soninke show moderately elevated scores, suggesting that Claude AI also has significant capabilities in these languages.

The human evaluation of translations carried out by the DNENF-LN and the AMALAN supported some of the results identified by automated scoring and contravened other results. Bambara and Soninke had highly positive evaluations by both human and automated scoring, while Kassonke's moderately elevated automated score was not confirmed by human evaluators. Fula, which from automated scoring seemed not to be meaningfully translated by Claude AI, was appreciated by the human evaluators as approximately equal to Bambara and Soninke. Songhay was also evaluated much more highly by humans than its automated scores would have suggested.

There is a correspondence in the results to the quantity of digital resources available in each language. Our team has created, worked with, or is aware of significant resources in Bambara, Soninke, Fula, and Songhay, all languages for which human evaluators gave good marks to the automated translations. A surprise in the results is the ability of Claude AI to produce any meaningful result in languages for which there are virtually no digital resources. The relatively high automated evaluation scores for Kassonke and low scores in human evaluation of meaningfulness of the text suggests a considerable ability to mimic language from very small samples without actually being able to produce intelligible text. The contrary lesson might be derived from the results with Fula, where poor automated scores suggests that Claude AI was able to translate the essence of the ideas covered by the text, causing it to forgo mimicry, producing a meaningful text with the vocabulary at its disposal.

**Conclusion**

Claude's performance in translating the languages of Mali, despite the lack of specific training, is, all limitations considered, impressive and highlights the immense potential of language models for underrepresented languages. This natural capacity demonstrates the interest in pushing further by enriching the corpora accessible to LLMs (large language models) in these languages.

Our investigation demonstrates that automated metrics may fail catastrophically, in some cases, in the assessment of translation quality for extremely low resourced languages. They are more reliable for languages with moderate resources, but it may be difficult to assess translation performance without human assessment. Humans provide more nuanced analysis of qualitative issues, particularly in terms of grammaticality and semantic fidelity that are needed to assess the ability of the model in the target language.

Cross-linguistic transfer and multilingual pre-training techniques have shown promising results, suggesting that the bar for the quantity of resources needed to obtain useful results has fallen significantly below what was supposed necessary earlier. It appears that relatively robust results for Bambara can be obtained despite its very small quantity of online data and curated training data. While the bar is not sufficiently low to enable translation to languages with the most limited resources, our results suggest that the effort to assemble even modest levels of resources for those languages may prove fruitful. It may also give additional incentive to work on further enhancing interlinguistic capabilities of LLMs so that the resources available and the capabilities of the LLMs may advance the state-of-art for low resource languages more quickly. We may not have reached the limits of interlinguistic capabilities of language models.

Page 7